\documentclass{article}
\usepackage{ijcai26}

\usepackage{times}
\usepackage{soul}
\usepackage{url}
\usepackage[hidelinks]{hyperref}
\usepackage[utf8]{inputenc}
\usepackage[small]{caption}
\usepackage{graphicx}
\usepackage{amsmath}
\usepackage{amsthm}
\usepackage{booktabs}
\usepackage{algorithm}
\usepackage{algorithmic}
\usepackage[switch]{lineno}

\usepackage{enumitem}   % 自定义列表样式
\usepackage{amssymb}

\usepackage{makecell}
\usepackage{array}
\usepackage{multirow}
\usepackage[normalem]{ulem}
\usepackage{subcaption}
\usepackage{tabularx}
\usepackage{color}
\usepackage{xspace} 
\usepackage[T1]{fontenc}
\usepackage{CJKutf8}
\usepackage{diagbox}
\usepackage[flushleft]{threeparttable}
\usepackage{marvosym}
\usepackage{amssymb}
\usepackage{fontawesome}
\usepackage{pifont}
\usepackage{xcolor}
\definecolor{mydarkgreen}{RGB}{0,100,0} 
\newcommand{\cmark}{\textcolor{mydarkgreen}{\ding{51}}} % Green checkmark
\usepackage{ulem}
\usepackage{xcolor}
\usepackage{color,xcolor}
\usepackage{tikz}

\usepackage[edges]{forest}
\definecolor{hidden-draw}{RGB}{0,0,0}
\definecolor{hidden-pink}{rgb}{0.98, 0.94, 0.75}
\definecolor{level0}{rgb}{0.67, 0.88, 0.69}
\definecolor{level1}{rgb}{0.98, 0.92, 0.84}
\definecolor{level2}{rgb}{0.8, 0.8, 1.0}
\definecolor{level3}{rgb}{1.0, 0.71, 0.76}
\definecolor{level4}{rgb}{0.49, 0.99, 0.0}

\title{A Comprehensive Survey of Deep Learning for Multivariate Time Series Forecasting: A Channel Strategy Perspective}

\author{Xiangfei Qiu$^{1}$\textsuperscript{\rm }, Hanyin Cheng$^{1}$\textsuperscript{\rm }, Xingjian Wu$^{1}$\textsuperscript{\rm }, Junkai Lu$^{1}$\textsuperscript{\rm },\\ Jilin Hu$^{1}$\textsuperscript{\rm }, Chenjuan Guo$^{1}$\textsuperscript{\rm }, Christian S. Jensen$^{2}$\textsuperscript{\rm }, and Bin Yang$^{1}$\textsuperscript{\rm }
    \affiliations
    \textsuperscript{\rm }$^1$School of Data Science and Engineering, East China Normal University, Shanghai, China \\ $^2$Department of Computer Science, Aalborg University, Aalborg, Denmark\\
    \emails{
    \{xfqiu, hycheng, xjwu, jklu\}@stu.ecnu.edu.cn}, \{jlhu, cjguo, byang\}@dase.ecnu.edu.cn, csj@cs.aau.dk}

\begin{document}

\maketitle

\begin{abstract}
    Multivariate Time Series Forecasting (MTSF) plays a crucial role across diverse fields, ranging from economics to energy to traffic. In recent years, deep learning has demonstrated outstanding performance in MTSF tasks. In MTSF, modeling the correlations among different channels is critical, as leveraging information from other related channels can significantly improve the prediction accuracy of a specific channel. This study systematically reviews the channel modeling strategies for time series and proposes a taxonomy organized into three hierarchical levels: the strategy perspective, the mechanism perspective, and the characteristic perspective. On this basis, we provide a structured analysis of these methods and conduct an in-depth examination of the advantages and limitations of different channel strategies. Finally, we summarize and discuss some future research directions to provide useful research guidance. Moreover, we maintain an up-to-date GitHub repository\footnote{\url{https://github.com/decisionintelligence/CS4TS}} which includes all the papers discussed in the survey.

% 多元时间序列预测（Multivariate Time Series Forecasting，MTSF）在经济、能源、人工智能运维（AIOps）以及交通等多个领域中发挥着至关重要的作用。近年来，深度学习在 MTSF 任务中展现了卓越的性能。在 MTSF 中，建模不同通道之间的相关性至关重要，因为利用其他相关通道的信息可以显著提升特定通道的预测精度。例如，在金融市场预测中，整合股票价格、交易量以及市场指数等数据，可以更准确地预测某一股票的未来走势。这些因素之间的相互关联为捕捉市场动态提供了更全面的视角。

% 本文系统性地回顾了时间序列的通道建模策略，并提出了分类法。在此基础上，我们对这些方法进行了结构化的梳理，并深入分析了不同通道策略的优势和局限性。最后，我们总结并探讨了五个未来研究方向，为相关领域的研究提供了宝贵的参考和指导。
\end{abstract}

\section{Introduction}
Multivariate time series forecasting (MTSF) is a fundamental yet challenging task in various domains, including economic, energy, and traffic~\cite{qiu2024tfb}. The ability to accurately predict future values of multiple interdependent channels (a.k.a., variables) over time is crucial for making informed decisions, optimizing resource allocation, and improving operational efficiency.

% \begin{figure}[t!]
%   \centering
%     \includegraphics[width=0.85\linewidth]{figs/intro.png}
%   \caption{The framework of our survey.}
%   \vspace{-3mm}
%   \label{intro}
% \end{figure}

\begin{figure}[t!]
  \centering
    \includegraphics[width=1\linewidth]{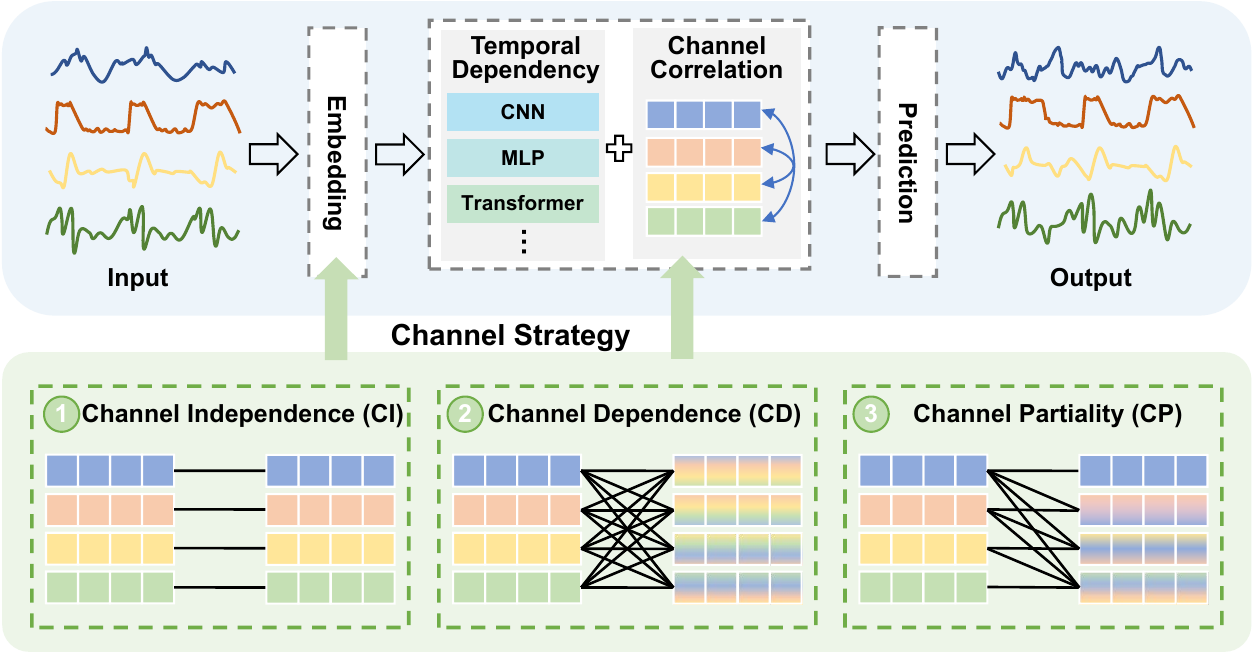}
    \vspace{-5mm}
  \caption{Channel strategy overview.}
      \vspace{-3mm}
  \label{channel strategy overview}
\end{figure}

In recent years, the rapid advancements in deep learning have significantly boosted the performance of MTSF. Researchers primarily model multivariate time series from the temporal and channel~(a.k.a., variable) dimensions. In terms of the temporal dimension, researchers have utilized various modules such as CNN~\cite{wu2022timesnet}, MLP~\cite{lincyclenet}, and Transformer~\cite{nie2022time} to capture the nonlinear and complex temporal dependencies within time series. In the channel dimension, researchers have designed different channel strategies to model the intricate correlations among channels~\cite{qiu2025duet,liu2023itransformer}. This is crucial in multivariate forecasting, as leveraging information from related channels can significantly improve the prediction accuracy of a specific channel. For example, in financial market forecasting, integrating data such as stock open price, trading volumes, and market indices can lead to more accurate predictions of a particular stock's return ratio. The interconnections among these factors provide a more comprehensive perspective for capturing market dynamics.

% 研究人员主要从时间维度和通道维度对多变量时间序列进行建模，其中，CNN、RNN 和 Transformer 等模型被广泛用于捕捉时间依赖关系。与此同时，为了有效建模变量间的相关性，研究人员提出了不同的通道策略。在多变量预测中，考虑变量之间的相关性至关重要，因为利用其他相关通道的信息可以显著提升特定通道的预测精度。例如，在金融市场预测中，将股价、交易量和市场指数等数据整合，可以更准确地预测某只股票的未来走势。这些变量之间的相互作用提供了更全面的视角，有助于更精准地捕捉市场动态。

% These deep learning approaches span various categories, including specific models like iTransformer~\cite{liu2023itransformer} and DUET~\cite{qiu2025duet}; foundation models such as Timer~\cite{Timer} and Chronos~\cite{chronos}; and plugin models like CCM~\cite{chen2024similarity} and LIFT~\cite{zhaorethinking}. 

% In recent years, the rapid advancements in deep learning have significantly boosted the performance of multivariate time series forecasting (MTSF), thanks to its remarkable capability to model complex temporal patterns and high-dimensional relationships. Compared to traditional statistical methods such as ARIMA~\cite{box1970distribution} and VAR~\cite{toda1994vector}, which are often limited in handling non-linear dependencies and long-term temporal structures, deep learning models have exhibited superior flexibility and accuracy. These deep learning approaches span various categories, including specific models like iTransformer~\cite{liu2023itransformer}, DUET~\cite{qiu2025duet}, and PatchTST~\cite{nie2022time}; foundation models such as MOIRAI~\cite{moiral}, UniTS~\cite{units}, and ROSE~\cite{rose}; and plugin models like CCM~\cite{chen2024similarity} and LIFT~\cite{zhaorethinking}. 

 Given the critical role of channel correlation in improving prediction accuracy, the selection of an appropriate channel strategy becomes a key design consideration. Overall, existing channel strategies---see Figure~\ref{channel strategy overview} can be categorized into three types: CI (Channel Independence) processing each channel independently without considering any potential interactions or correlations among them~\cite{nie2022time,lincyclenet}; CD (Channel Dependence) treating all channels as a unified entity, assuming they are interrelated and dependent on each other~\cite{zhang2022crossformer,liu2023itransformer}; and CP (Channel Partiality) meaning that each channel maintains some degree of independence while simultaneously being influenced by some other related channels~\cite{MCformer,qiu2025duet}. 
 Each strategy reflects a unique perspective on how to model the inter-channel dependencies, leading to diverse modeling architectures and applications.

 % 本文首先简要介绍了多通道时间序列预测（MTSF）任务，并提出了一种新的分类方法，将其划分为三个层次。第一部分从策略视角出发，系统阐述了三种通道策略的定义及其代表性方法，为研究人员提供了理解这些策略的基础框架；第二部分从机制视角进一步探讨了各方法如何实现通道策略的具体机制，进行分类总结，以帮助研究人员深入理解不同通道策略背后的实现原理；最后，第三部分从特性视角出发，重点分析了通道策略在建模变量间相关性时所考虑的不同特性。接着，本文对CI、CD和CP三种通道策略的优缺点进行了详细对比，为研究人员提供了有价值的见解。最后，我们讨论了未来的研究方向，为该领域的进一步发展提供了有益的指导，并总结了本研究的主要贡献。
 
% presenting a detailed taxonomy and critical analysis of deep learning approaches for MTSF, with an emphasis on how different channel strategies impact the design and performance of these models. 

\textcolor{black}{While there are several surveys on MTSF~\cite{DBLP:conf/ijcai/WenZZCMY023,wang2024deep}, they often lack a comprehensive discussion on the role of channel strategies in multivariate settings. This study aims to bridge this gap by summarizing the main developments of channel strategies of MTSF. We first briefly introduce the MTSF task and propose a new taxonomy organized into three hierarchical levels. Starting with the strategy perspective, we systematically introduce the definitions of three channel strategies and their representative methods, providing researchers with a foundational framework to understand these strategies. Next, from the mechanism perspective, we further explore how each method implements the specific mechanisms of channel strategies, categorizing and summarizing them to help researchers gain a deeper understanding of the underlying implementation of different channel strategies. Finally, from the characteristic perspective, we focus on the different characteristics considered by channel strategies when modeling the correlations among channels. Following this, we provide a detailed comparison of the advantages and limitations of the three-channel strategies, offering valuable insights for researchers. In conclusion, we discuss future research directions that will provide useful guidance for the further development of the field. To the best of our knowledge, this is the first work to comprehensively and systematically review the key developments of deep learning methods for MTSF through the lens of channel strategy.} In summary, the main contributions of this survey include:
\begin{itemize}
\item \textbf{Comprehensive and up-to-date survey:} We provide an in-depth review of state-of-the-art deep learning models for MTSF, highlighting their use of channel strategies. 

% We examine key model components, including learning paradigms and architecture.
\item \textbf{Novel channel-perspective taxonomy:} We introduce a structured taxonomy of channel strategies for deep learning-based MTSF, offering a comprehensive analysis of their strengths and limitations.
\item \textbf{Future research opportunities :} We discuss and highlight future avenues for enhancing MTSF through diverse channel strategies, urging researchers to delve deeper into this area.
\end{itemize}

\section{Preliminaries}
\noindent
~~~\textbf{Time Series:} 
A time series  $X \in \mathbb{R}^{T\times N}$ is a time-oriented sequence of N-dimensional time points,  where $T$ is the number of timestamps, and $N$ is the number of channels. 
% For convenience, we separate dimensions with commas. Specifically, we denote $X_{i,j} \in \mathbb{R}$ as the $j$-th channel at the $i$-th timestamp, $X_{n,:}\in \mathbb{R}^N$ as the time series of $n$-th timestamp, where $n=1,\cdots,T$.

\textbf{Multivariate Time Series Forecasting:} 
Given a historical multivariate time series $X \in \mathbb{R}^{T\times N}$ of $T$ time points, MTSF aims to predict the next $F$ time points, i.e., $Y\in\mathbb{R}^{F\times N}$, where $F$ is called the forecasting horizon.

\begin{table*}[t!]
  \centering
  \caption{A taxonomy of channel strategy in multivariate time series forecasting. }
    \vspace{-2mm}
  \label{Channel Strategy}
  \resizebox{0.95\linewidth}{!}{
\begin{tabular}{c|c|cccccc|c|cccc}
\toprule
\multirow{2}[1]{*}{Strategy}    & \multirow{2}[1]{*}{Mechanism}   & \multicolumn{6}{c|}{Characteristic} & \multirow{2}[1]{*}{Method}   & \multirow{2}[1]{*}{Paradigm}  &  \multirow{2}[1]{*}{Venue} & \multirow{2}[1]{*}{Year}  & \multirow{2}[1]{*}{Code}  \\
\cmidrule{3-8} & &Asym.& Lag.& Pol.& Gw. & Dyn. & Ms.&  \\
\midrule
\multirow{8}{*}{\rotatebox{90}{CI}}  & - &- & - & -&- & - &- & PatchTST~\cite{nie2022time} & Specific  & ICLR & 2023 & \color{blue}\href{https://github.com/yuqinie98/patchtst}{PatchTST}  \\ 
%   & - & - &- & - & -&- & - & Triformer~\cite{Triformer} & Specific Model & IJCAI & 2022 & \color{blue}\href{https://github.com/razvanc92/triformer}{Triformer}   \\
  %  & - &- & - & -&-  & - &-  & PDF~\cite{PDFliu} & Specific Model & ICLR & 2024 & \color{blue}\href{https://github.com/Hank0626/PDF}{PDF}   \\
%   & - &- & - & -&- & - &- & SparseTSF~\cite{lin2024sparsetsf}  & Specific Model & ICML & 2024 & \color{blue}\href{https://github.com/lss-1138/SparseTSF}{SparseTSF}   \\
  & - &- & - & -&-  & - &-  & CycleNet~\cite{lincyclenet} & Specific  & NIPS & 2024 & \color{blue}\href{https://github.com/ACAT-SCUT/CycleNet}{CycleNet}   \\
  & - &- & - & -&- & - &- & DLinear~\cite{zeng2023transformers}  & Specific  & AAAI & 2023 & \color{blue}\href{https://github.com/cure-lab/LTSF-Linear}{DLinear}   \\
 % Channel Independent   & MLP-based &-  & NLinear~\cite{zeng2023transformers} & Specific  & AAAI & 2023 & \color{blue}\href{https://github.com/cure-lab/LTSF-Linear}{NLinear}   \\
 % Channel Independent  & MLP-based &- & TimeMixer~\cite{wang2024timemixer} & Specific  & ICLR & 2024 & \color{blue}\href{https://github.com/kwuking/TimeMixer}{TimeMixer}   \\
%  & - & - &- & - & -&- & - & TimeMixer~\cite{wang2024timemixer}  & Specific Model & ICLR & 2024 & \color{blue}\href{https://github.com/kwuking/TimeMixer}{TimeMixer}   \\
% & - & - &- & - & -&- & - & Pathformer~\cite{chen2024pathformer} & Specific Model   & ICLR & 2024 & \color{blue}\href{https://github.com/decisionintelligence/pathformer}{Pathformer}   \\
 & - &- & - & -&- & - &- & Timer~\cite{Timer} & Foundation  & ICML & 2024 & \color{blue}\href{https://github.com/thuml/Large-Time-Series-Model}{Timer}   \\
  & - &- & - & -&- & - &- & One-Fits-All~\cite{GPT2} & Foundation  & NIPS & 2023 & \color{blue}\href{https://github.com/DAMO-DI-ML/NeurIPS2023-One-Fits-All}{One-Fits-All}   \\
   & - &- & - & -&- & - &- & $S^2$IP-LLM~\cite{S2IP-LLM} & Foundation  & ICML & 2024 & \color{blue}\href{https://github.com/KimMeen/Time-LLM}{$S^2$IP-LLM}   \\
    & - &- & - & -&- & - &- & TEMPO~\cite{tempo} & Foundation  & ICLR & 2024 & \color{blue}\href{https://github.com/DC-research/TEMPO}{TEMPO}   \\
 % Channel Independent & Transformer-based &- & TimesFM~\cite{timesfm} & Foundation  & ICML & 2024 & \color{blue}\href{https://github.com/google-research/timesfm/}{TimesFM}   \\
 % & - &- & - & -&-  & - &-  & Chronos~\cite{chronos}  & Foundation  & ICML & 2024 & \color{blue}\href{https://github.com/amazon-science/chronos-forecasting}{Chronos}   \\
 %  & - &- & - & -&-  & - &-  & Time-MoE~\cite{time-moe} & Foundation  & arXiv & 2024 & \color{blue}\href{https://github.com/Time-MoE/Time-MoE}{Time-MoE}   \\
 %     & - &- & - & -&- & - &- & ROSE~\cite{rose} & Foundation  & arXiv & 2024 &  - \\
 % Channel Independent   & Transformer-based &- & GPT4TS~\cite{gpt4ts} & Foundation  & NIPS & 2023 & \color{blue}\href{https://github.com/DAMO-DI-ML/NeurIPS2023-One-Fits-All}{GPT4TS} \\
 % & - &- & - & -&- & - &- &LLM4TS~\cite{llm4ts}    & Foundation  & NIPS & 2023 & \color{blue}\href{https://github.com/liaoyuhua/LLM4TS}{LLM4TS} \\
   & - &- & - & -&- & - &- & Time-LLM~\cite{time-llm}  & Foundation  & ICLR & 2024 & \color{blue}\href{https://github.com/KimMeen/Time-LLM}{Time-LLM} \\
 % Channel Independent  & Transformer-based &- & $\textbf{S}^2$IP-LLM~\cite{s2ip} & Foundation  & ICML & 2024 & \color{blue}\href{https://github.com/panzijie825/S2IP-LLM}{$\textbf{S}^2$IP-LLM} \\
   % & - &- & - & -&- & - &- & Tempo~\cite{tempo} & Foundation  & ICLR & 2024 & \color{blue}\href{https://github.com/DC-research/TEMPO}{Tempo} \\ 
    & - &- & - & -&- & - &- & RevIN~\cite{RevIN} & Plugin  & ICLR & 2021 & \color{blue}\href{https://github.com/ts-kim/RevIN}{RevIN} \\ \hline
 % Channel Independent & Foundation Model (LLM)  & Transformer-based &- & Autotimes~\cite{autotimes} & NIPS & 2024 & \color{blue}\href{https://github.com/thuml/AutoTimes}{Autotimes} \\
\multirow{23}{*}{\rotatebox{90}{CD}} & CNN-based & \cmark & - & - & -&- & - &Informer~\cite{zhou2021informer}   & Specific  & AAAI & 2021 & \color{blue}\href{https://github.com/zhouhaoyi/Informer2020}{Informer}   \\
  & CNN-based & \cmark & - & - & -&- & - &Autoformer~\cite{wu2021autoformer} & Specific  & NIPS  & 2021 & \color{blue}\href{https://github.com/thuml/Autoformer}{Autoformer}   \\
 & CNN-based &\cmark& - & -&- & - &- & FEDformer~\cite{zhou2022fedformer} & Specific  & ICML & 2022 & \color{blue}\href{https://github.com/MAZiqing/FEDformer}{FEDformer}   \\
  & CNN-based & \cmark &- & - & -&- & - &TimesNet~\cite{wu2022timesnet} & Specific  & ICLR & 2023 & \color{blue}\href{https://github.com/thuml/TimesNet}{TimesNet}   \\
%  & MLP-based & \cmark &- & - & -&- & - & TSMixer~\cite{chen2023tsmixer} & Specific  & TMLR & 2023 & \color{blue}\href{https://github.com/google-research/google-research/tree/master/tsmixer}{TSMixer}   \\
  & MLP-based & \cmark &- & - & -&- & - & TSMixer~\cite{ekambaram2023tsmixer} & Specific  & KDD & 2023 & \color{blue}\href{https://github.com/ditschuk/pytorch-tsmixer}{TSMixer}   \\
   & MLP-based &\cmark & - &- & - & -&- & TTM~\cite{ttm}   & Foundation  & NIPS  & 2024 & \color{blue}\href{https://github.com/ibm-granite/granite-tsfm}{TTM}   \\ 
 & Transformer-based & \cmark &- & - & -&\cmark & - & iTransformer~\cite{liu2023itransformer} & Specific  & ICLR & 2024 & \color{blue}\href{https://github.com/thuml/iTransformer}{iTransformer}   \\
  % Channel Independent & Multimodal Model  & Transformer-based & ChatTime~\cite{ChatTime} & AAAI & 2025 & \color{blue}\href{https://github.com/forestsking/chattime}{ChatTime}   \\
  % - & Multimodal Model  & - & Hybrid-MMF~\cite{kim2024multi} & arXiv & 2024 & \color{blue}\href{https://github.com/Rose-STL-Lab/Multimodal_Forecasting}{Hybrid-MMF}   \\
  & Transformer-based & \cmark  &- & - & -&\cmark & - & Crossformer~\cite{zhang2022crossformer} & Specific  & ICLR & 2023 & \color{blue}\href{https://github.com/Thinklab-SJTU/Crossformer}{Crossformer}   \\
 % & Transformer-based & - &- & - & -&- & - & SAMformer~\cite{ilbert2024samformer}  & Specific  & ICML & 2024 & \color{blue}\href{https://github.com/romilbert/samformer}{SAMformer}   \\
 & Transformer-based & \cmark & \cmark &- & - & -&- & VCformer~\cite{vcformer}   & Specific  & IJCAI & 2024 & \color{blue}\href{https://github.com/CSyyn/VCformer}{VCformer}   \\
   & Transformer-based &\cmark & \cmark &- & - & \cmark&- & MOIRAI~\cite{MOIRAI}  & Foundation  & ICML  & 2024 & \color{blue}\href{https://github.com/SalesforceAIResearch/uni2ts}{MOIRAI}   \\
  & Transformer-based &\cmark & - &- & - &\cmark&- & UniTS~\cite{units}   & Foundation  & NIPS  & 2024 & \color{blue}\href{https://github.com/mims-harvard/UniTS}{UniTS}   \\
    & Transformer-based &\cmark& - &- & -& \cmark& - & TQNet~\cite{lin2025tqn}  & Specific  & ICML & 2025 &  \color{blue}\href{https://github.com/ACAT-SCUT/TQNet}{TQNet}   \\
 & GNN-based &  & - &- & - & \cmark&- & GTS~\cite{gts}  & Specific  &ICLR   & 2021 & \color{blue}\href{https://github.com/chaoshangcs/GTS}{GTS}   \\
   & GNN-based &\cmark & - &- & - & -&\cmark & MSGNet~\cite{MSGNet} & Specific  & AAAI & 2024 & \color{blue}\href{https://github.com/YoZhibo/MSGNet}{MSGNet}   \\
 & GNN-based & - &\cmark &- & - & -&- & FourierGNN~\cite{FourierGNN}  & Specific  & NIPS  & 2023 & \color{blue}\href{https://github.com/aikunyi/FourierGNN}{FourierGNN}   \\
& GNN-based & - &\cmark&- & - & -&- & FC-STGNN~\cite{FC-STGNN}  & Specific  & AAAI & 2024 & \color{blue}\href{https://github.com/Frank-Wang-oss/FCSTGNN}{FC-STGNN }   \\
 & GNN-based &\cmark& - &- & - &\cmark&- & TPGNN~\cite{TPGNN}  & Specific  & NIPS  & 2022 & \color{blue}\href{https://github.com/zyplanet/TPGNN}{TPGNN}   \\
 & GNN-based &\cmark& - &- & - & \cmark&\cmark & ESG~\cite{ESG}  & Specific  & KDD  & 2022 & \color{blue}\href{https://github.com/LiuZH-19/ESG}{ESG}   \\
 & GNN-based &\cmark& - &- & - &\cmark&\cmark & EnhanceNet~\cite{EnhanceNet}   & Plugin  & ICDE  & 2021 & \color{blue}\href{https://github.com/razvanc92/EnhanceNet}{EnhanceNet} \\
& Others &\cmark& - &- & - &-&-& SOFTS~\cite{SOFTS}   & Specific  & NIPS  & 2024 & \color{blue}\href{https://github.com/Secilia-Cxy/SOFTS}{SOFTS} \\
& Others &\cmark& - &- & - &\cmark& - & C-LoRA~\cite{C-LoRA}   & Plugin  & CIKM  & 2024 & \color{blue}\href{https://github.com/tongnie/C-LoRA}{C-LoRA} 
    \\ \hline
 \multirow{12}{*}{\rotatebox{90}{CP}}   
 & CNN-based&\cmark &- & - & -&- & - &ModernTCN~\cite{donghao2024moderntcn}   & Specific  & ICLR & 2024 & \color{blue}\href{https://github.com/luodhhh/ModernTCN}{ModernTCN}   \\
  & Transformer-based &\cmark& - &- & \cmark& - & - & DUET~\cite{qiu2025duet}  & Specific  & KDD & 2025 &  \color{blue}\href{https://github.com/decisionintelligence/DUET}{DUET}   \\
  & Transformer-based & \cmark & - & - & -&\cmark& - &MCformer~\cite{MCformer}  & Specific  &  IITJ$^{*}$ & 2024 & -  \\
    & Transformer-based&\cmark&- & - &\cmark&\cmark & - & DGCformer~\cite{liu2024dgcformer}  & Specific  & arXiv & 2024 & -  \\
  & Transformer-based & - &- & - & - &\cmark & - & CM~\cite{lee2024partial}  & Plugin  & NIPS & 2024 & -  \\
 & GNN-based & \cmark & - &- & - & -&- & MTGNN~\cite{MTGNN}  & Specific  & KDD  & 2020 & \color{blue}\href{https://github.com/nnzhan/MTGNN}{MTGNN}   \\
 & GNN-based & \cmark& - &\cmark& - & -&- & CrossGNN~\cite{CrossGNN}  & Specific  & NIPS  & 2023 & \color{blue}\href{https://github.com/hqh0728/CrossGNN}{CrossGNN}   \\
  & GNN-based & \cmark& - & & - &\cmark&- & WaveForM~\cite{WaveForM}   & Specific  & AAAI  & 2023 & \color{blue}\href{https://github.com/alanyoungCN/WaveForM}{WaveForM}   \\
  & GNN-based & - & - &- & - &\cmark&- & MTSF-DG~\cite{zhao2023multiple}   & Specific  & VLDB  & 2023 & \color{blue}\href{https://github.com/decisionintelligence/MTSF-DG}{MTSF-DG}   \\
 & GNN-based & \cmark & - &- & \cmark & -&- & ReMo~\cite{ReMo}    & Specific  & IJCAI  & 2023 & -   \\
 & GNN-based & -  & \cmark &- & \cmark & \cmark &- & TimeFilter~\cite{hu2025timefilter}    & Specific  & ICML  & 2025 & \color{blue}\href{https://github.com/TROUBADOUR000/TimeFilter}{TimeFilter}    \\
 % & GNN-based &\cmark& - &- &\cmark & \cmark&\cmark & Ada-MSHyper~\cite{Ada-MSHyper}     & Specific  & NIPS  & 2024 & \color{blue}\href{https://github.com/shangzongjiang/Ada-MSHyper}{Ada-MSHyper}   \\
  &Others& \cmark& \cmark & -&- & - & - & LIFT~\cite{zhaorethinking}  & Plugin  & ICLR & 2024 & \color{blue}\href{https://github.com/SJTU-DMTai/LIFT}{LIFT}   \\
  & Others & \cmark &- & - & \cmark&- & - & CCM~\cite{chen2024similarity}  & Plugin  & NIPS & 2024 & \color{blue}\href{https://github.com/Graph-and-Geometric-Learning/TimeSeriesCCM}{CCM}   \\ 
 
 \bottomrule
 \multicolumn{13}{l}{Asym.: Asymmetry, Lag.: Lagginess, Pol.: Polarity, Gw.: group-wise, Dyn.: Dynamism, and Ms.: Multi-scale. We will discuss them in \textcolor{black}{Section}~\ref{Characteristics Perspective}.}\\
\multicolumn{13}{l}{Since the CI models do not consider the correlations among channels, the corresponding positions for mechanism and characteristic are marked as ``-".} \\
\multicolumn{13}{l}{For the plugin model, its mechanism is complex, so we exclude it, marking the corresponding position as "-"; IITJ$^{*}$: IEEE Internet Things J.}
\end{tabular}
}
\vspace{-3mm}
\end{table*}

\section{Taxonomy of Channel Strategies in MTSF}
The taxonomy presented in Table~\ref{Channel Strategy} provides a structured classification to enhance the understanding of channel strategies in MTSF. It is organized into three hierarchical levels, starting with the strategy perspective, followed by the mechanism perspective, and finally the characteristic perspective. 

\subsection{Strategy Perspective}
A channel strategy refers to the approach employed to process, integrate, or utilize information from multiple input channels. As illustrated in Figure~\ref{channel strategy overview}, the explored strategies can be broadly categorized as follows. \textcolor{black}{We will discuss the pros and cons of these strategies in Section~\ref{Comparison within the Taxonomy}.}

% \subsubsection{Channel Independent}
% \noindent
\textbf{Channel Independence (CI):} 
The CI strategy treats each channel independently, without considering any potential interactions or correlations among channels. Each channel is processed as a separate input, and no shared information or dependencies are utilized. The representative method PatchTST~\cite{nie2022time} employs CI and demonstrates outstanding performance in MTSF. This design significantly reduces model complexity, enabling faster inference while mitigating the risk of overfitting caused by noise or spurious correlations among channels. Furthermore, the CI strategy offers flexibility, as the addition of new channels does not require changes to the model architecture, allowing it to seamlessly adapt to evolving datasets. These advantages have made the CI strategy increasingly popular in recent research, 
contributing to improved forecasting performance~\cite{lincyclenet,zeng2023transformers}.

% \begin{figure*}[t!]
%   \centering
%     \includegraphics[width=0.8\linewidth]{figs/corr-overview.pdf}
%   \caption{Channel strategy overview.}
%     \vspace{-3mm}
%   \label{channel strategy overview}
% \end{figure*}

% \subsubsection{Channel Dependence}
% \noindent
\textbf{Channel Dependence (CD):} 
The CD strategy assumes that all channels in a multivariate time series are inherently correlated and interdependent, treating them as a unified entity during the forecasting process.
\textcolor{black}{
Based on the phases when inter-channel interactions are learned, the existing CD methods can be divided into two categories: I) \textcolor{black}{\textbf{Embedding fusion:}} These models fuse data from different channels when obtaining their time series embedding representations. For example, Informer~\cite{zhou2021informer}, Autoformer~\cite{wu2021autoformer}, and TimesNet~\cite{wu2022timesnet} use 1D or 2D convolutions to extract temporal representations. In the convolutional operation, each convolutional kernel first performs a sliding convolution within each input channel to obtain the corresponding feature maps. These feature maps for all channels are then weighted and combined, capturing the dependencies among the channels.
II) \textbf{Explicit correlation:} These models often design specialized modules to explicitly model channel correlations, facilitating more structured channel modeling based on the acquired time series embedding representations. 
Representative algorithms include iTransformer~\cite{liu2023itransformer} and TSMixer~\cite{ekambaram2023tsmixer}. iTransformer adopts a self-attention module among channels, treating independent time series as tokens and capturing multivariate correlations using the self-attention mechanism. In contrast, TSMixer uses an MLP module among channels to capture the intricate correlations among channels, with these correlations represented by multi-level features extracted through fully connected layers.}

% Examples include MLP-based models (TSMixer~\cite{ekambaram2023tsmixer}, TTM~\cite{ttm}), Transformer-based models (iTransformer~\cite{liu2023itransformer}, Crossformer~\cite{zhang2022crossformer}), and GNN-based models (GTS~\cite{gts}, MSGNet~\cite{MSGNet}, TPGNN~\cite{TPGNN}), etc. The channel interactions based on different mechanisms will be explained in detail in Section~\ref{Mechanism Perspective}.}

\begin{figure}[t!]
  \centering
    \includegraphics[width=1\linewidth]{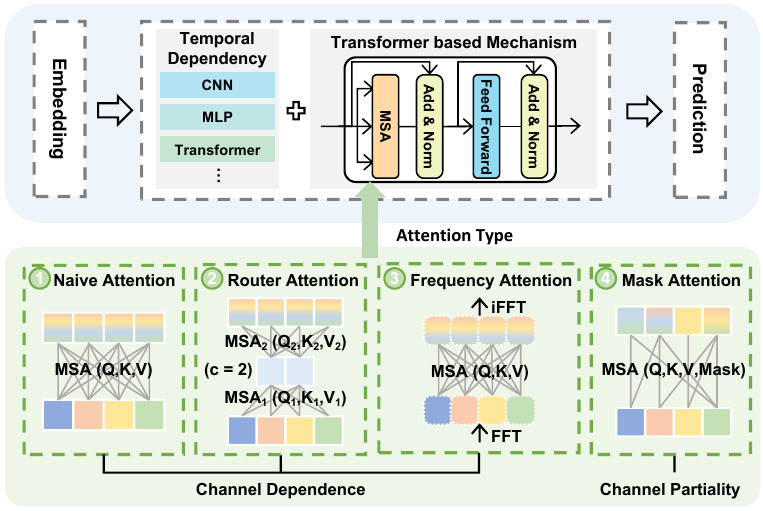}
  \caption{Transformer-based mechanism for channel strategy.}
    \vspace{-4mm}
  \label{Transformer}
\end{figure}

\textbf{Channel Partiality (CP):} 
 The CP strategy strikes a balance between CI and CD, allowing each channel to retain a degree of independence while simultaneously interacting with other related channels. This approach emphasizes a hybrid state where channels selectively interact and exhibit partial correlations. Based on whether the number of related channels for each channel is fixed or dynamic, the existing CP methods can be divided into two categories:
 % \textcolor{black}{Based on whether the number of related channels for each channel is fixed or dynamic, CP can be further refined into CHP (Channel Hard Partiality) and CSP (Channel Soft Partiality). CHP means that the number of related channels for each channel is fixed and predetermined, meaning the set of associated channels remains constant over time. CSP refers to that the number of related channels for each channel is dynamically and can change over time, allowing for more flexibility in adapting to varying scenarios.}
% a method in which each channel interacts dynamically and flexibly with its associated channels. 
% CHP models, such as \textcolor{red}{DGCformer}~\cite{liu2024dgcformer}, use graph clustering modules to group channels with significant similarities into the same cluster. Within each cluster, the CD strategy is applied to model interdependencies, while the CI strategy is used across clusters to maintain computational efficiency and avoid overfitting.
% CHP 模型，往往在讨论通道间关系时，对通道进行较为死板的划分，如 MTGNN~\cite{liu2024dgcformer}，将通道间关系建模为K-正则图，即每个通道都与K个通道应用 CD 策略来建模通道间的相互依赖，而与其余N-K个通道应用 CI 策略。与之类似的MCformer中每个通道也只与K个通道进行混合，而对其余N-K个通道保持 CI 策略以保持计算效率并避免过拟合。
\textcolor{black}{I) \textbf{Fixed partial channels:} These models fix the number of related channels for each channel, which means the set of associated channels remains constant over time. For example, in MTGNN~\cite{MTGNN}, the channel relationships are modeled as a K-regular graph, where each channel interacts with \(K\) other channels using the CD strategy to model interdependencies, while the remaining channels interact through the CI strategy. Similarly, in MCformer~\cite{MCformer}, each channel interacts with only \(K\) other channels, maintaining the CI strategy with the rest to ensure computational efficiency and prevent overfitting.
II) \textbf{Dynamic partial channels:} These models allow the number of related channels for each channel to be dynamic, changing over time and providing greater flexibility to adapt to varying scenarios. For instance, DUET~\cite{qiu2025duet} calculates channel similarity using metric learning in the frequency domain and then sparsifies the result. This creates a mask matrix, which is integrated into the attention mechanism of the fusion module, ensuring that each channel interacts only with relevant channels, reducing interference from noisy ones. Another example, CCM~\cite{chen2024similarity}, dynamically clusters channels based on their intrinsic similarities. To effectively capture the underlying time series patterns within these clusters, CCM utilizes a cluster-aware feed-forward mechanism, enabling tailored management and processing for each cluster.}

\subsection{Mechanism Perspective}
\label{Mechanism Perspective}
This section presents mainly the various mechanisms designed to model the relationships among channels.
% \subsubsection{Transformer-based Mechanism}

~~\textbf{Transformer-based:}  
% As shown in Figure~\ref{Transformer}, existing channel strategies based on the Transformer mechanism can be categorized into the following types. I) \textbf{Naive Attention}: These approaches treat time series segments (patches) or the entire sequence of each channel as individual tokens, directly applying attention mechanisms to model inter-channel relationships. II) \textbf{Router Attention}: These approaches introduce a Router Mechanism for Naive Attention, which uses a small fixed number of ``routers" (\(c\)) to gather information from all channels and redistribute it. This reduces the complexity from \(O(N^2)\) to \(O(2cN) = O(N)\). III) \textbf{Mask Attention}: These approaches generate mask matrices for Naive Attention, allowing each channel to focus on those beneficial for downstream prediction tasks, while mitigating the impact of noisy or irrelevant channels. IV) \textbf{Frequency Attention}: These approaches transform the time series data into the frequency domain and then employ Naive Attention to model inter-channel relationships. 
In recent years, Transformer has been widely applied to MTSF tasks, leveraging its powerful global modeling capability to effectively capture complex temporal dependencies and channel interactions. As shown in Figure~\ref{Transformer}, existing channel strategies based on the attention mechanism can be categorized into the following types. I) \textbf{Naive Attention:} These approaches all adopt the CD strategy, treating time series segments (patches) or the entire sequence of each channel as individual tokens, and directly applying attention mechanisms to model channel correlations. For instance, CARD~\cite{CARD} and iTransformer~\cite{liu2023itransformer} represent the patches and series of each channel as independent tokens, respectively, and explicitly capture channel correlation using attention mechanisms. II) \textbf{Router Attention:} When the number of channels (\(N\)) is large, the computational complexity of channel attention reaches \(O(N^2)\), resulting in high computational costs. To address this, some methods propose optimization strategies to mitigate the computational complexity caused by the CD strategy. For example, Crossformer~\cite{zhang2022crossformer} introduces a Router Mechanism for Naive Attention, which uses a small fixed number of \textcolor{black}{$c$ ``routers" ($c \ll N$)} to gather information from all channels and redistribute it. This reduces the complexity to \(O(2cN) = O(N)\). This mechanism effectively balances the modeling of channel correlation and computational efficiency. III) \textbf{Frequency Attention:} Some CD methods suggest that frequency-domain information is more effective for capturing inter-channel dependencies than time-domain information. For example, FECAM~\cite{FECAM} transforms the time series data into the frequency domain and then employs Naive Attention to model inter-channel relationships in this domain. IV) \textbf{Mask Attention:} In naive attention, each channel calculates attention scores with all channels, which can be negatively affected by irrelevant channels. To mitigate this, Mask Attention provides an approach to avoid irrelevant noise by constructing a CP strategy. For example, DUET~\cite{qiu2025duet} generates mask matrices for Naive Attention, allowing each channel to focus on those beneficial for downstream prediction tasks, while mitigating the impact of noisy or irrelevant channels. This approach explicitly constrains the attention computation, improving the accuracy of channel correlation modeling.

\begin{figure}[t!]
  \centering
    \includegraphics[width=1\linewidth]{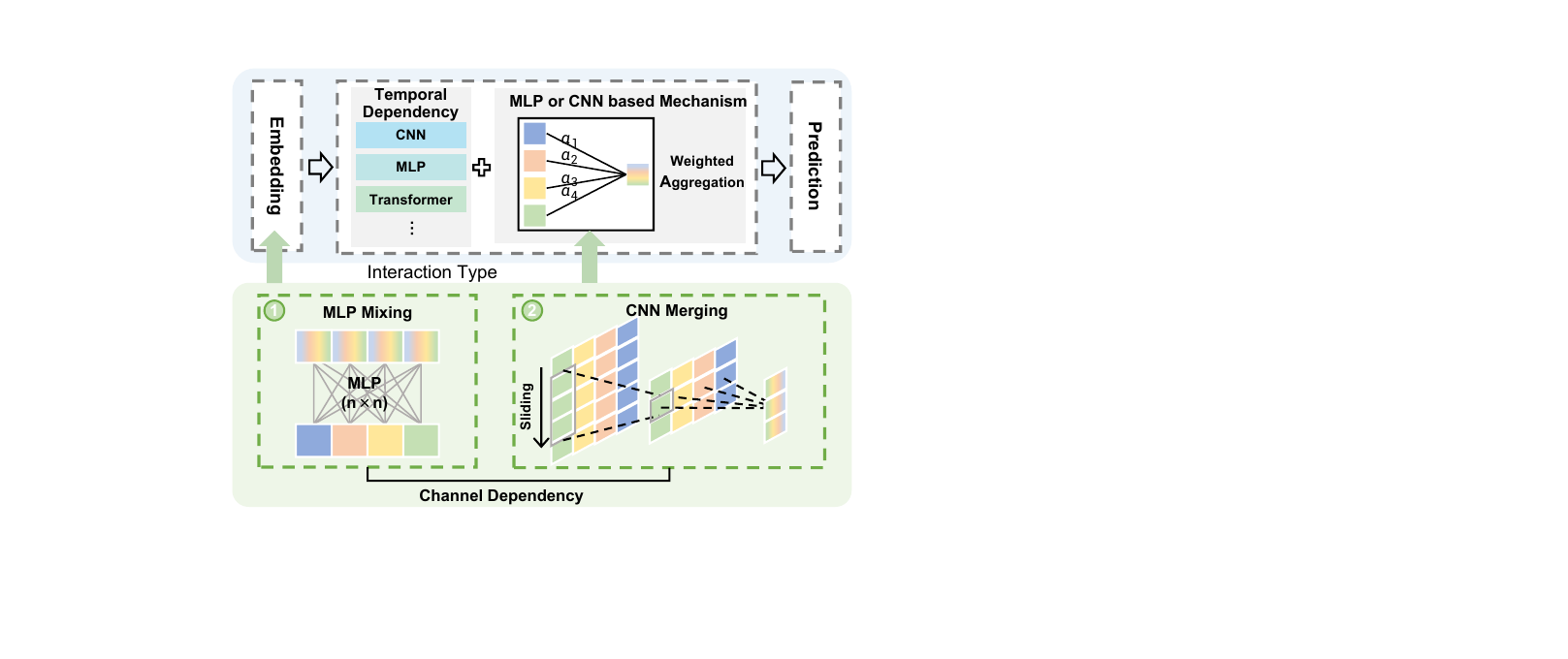}
  \caption{MLP, CNN-based mechanism for channel strategy.}
    \vspace{-4mm}
  \label{CNN}
\end{figure}

% \noindent
\textbf{MLP-based:} 
% 多层感知机（MLP）作为骨干网络，具有强大的特征学习能力，通过多层非线性变换，将输入空间映射到输出空间，从而学习数据的复杂特征表示。因为其结构简单且易于实现，在多变量时间序列预测中得到广泛应用。
The Multilayer Perceptron (MLP), as a backbone network, possesses powerful feature learning capabilities according to the universal approximation theorem. Existing MLP-based models use \textbf{MLP Mixing} in a CD manner, to capture the intricate correlations among channels, with these correlations represented by multi-level features extracted through fully connected layers---see Figure~\ref{CNN}. From the perspective of channel strategies, models in the MLP Mixing category, such as TSMixer~\cite{ekambaram2023tsmixer} and Tiny-TTM~\cite{ttm}, employ this approach to efficiently \textcolor{black}{capture correlations among all channels}, achieving strong performance with low computational cost, and all fall under the CD strategy.

\textbf{CNN-based:} 
A Convolutional Neural Network (CNN) is a deep learning model that utilizes convolutional layers to extract local features from data.
% As illustrated in Figure~\ref{CNN}, the explored CNN-based approaches can be broadly categorized as follows:
% \textbf{I) Merging}: These approaches primarily use 1D convolution with a sliding operation along the temporal dimension in the initial feature extraction layers. By treating different time series channels as distinct convolutional input channels, the models weight and merge the information from various channels during the convolution process, enabling interactions between time series channels.
% \textbf{II) Convolution}: These approaches directly apply convolution operations along the channel dimension, facilitating information interaction between channels within a local scope. Within the same convolution window, channels interact with each other through the convolution kernel, while channels that cannot be assigned to the same window remain independent of each other. 
As illustrated in Figure~\ref{CNN}, the explored CNN-based approaches use \textbf{CNN Merging} for channel modeling, such as Informer~\cite{zhou2021informer}, Autoformer~\cite{wu2021autoformer}, and FEDformer~\cite{zhou2022fedformer}, use 1D convolution with a sliding operation along the temporal dimension in the initial feature extraction layers. These models treat different channels as distinct inputs to the convolution, whose features are then weighted and merged during the convolution process, enabling inter-channel interactions. Although TimesNet~\cite{wu2022timesnet} employs 2D convolutions, it folds the temporal dimension into a 2D format, with channels still serving as independent input for weighted merging via convolution. Such models are all under the CD strategy.

% II)~\textbf{Convolution:} Given the slight spatial dependence among channels, ModerTCN~\cite{donghao2024moderntcn} directly \textcolor{black}{applies convolution operations to facilitate information interaction among channels within local scopes}. Within the same convolution window, channels interact with each other in a CD manner through the convolution kernel, while channels that cannot be assigned to the same window remain independent of each other. This results in an efficient method for CP modeling.

% Although TimesNet employs 2D convolutions, it folds the temporal dimension into a 2D format, with variable channels still serving as input channels for weighted merging via convolution.

% \subsubsection{GNN-based Mechanism}
% \noindent
% From the perspective of channel strategies, the aforementioned GNN-based methods can be classified into dense and sparse graphs. In dense graphs, each node is connected to almost all other nodes, so methods based on dense graphs typically adopt the CD strategy. In contrast, in sparse graphs, only necessary edges exist, with most nodes remaining independent, so methods based on sparse graphs are more likely to adopt the CP strategy.

% 通过将窗口内每个通道视为节点，通道间的相关性视为边，多变量时间序列可以转换为图结构数据。基于图的方法可以分为稠密图和稀疏图。在稠密图中，每个节点几乎与所有其他节点之间都有边，这样的边常常表示相关性的强弱程度或相关影响的存在概率。如GTS，FourierGNN等。基于稠密图的方法通常属于CD策略。而在稀疏图中，仅存在必要的边连接，大部分节点保持独立，如MTGNN为每个节点保留了K条边，构建了稀疏的K正则图。与之不同的MTSF-DG通过预设的阈值，过滤掉低概率的边，以稀疏化邻接矩阵。这样的基于稀疏图的方法则更倾向于CP策略。

\textbf{GNN-based:}
% By treating each channel within the \textcolor{black}{look-back window} as a node and the correlations between channels as edges, multivariate time series can be transformed into graph-based data. 
\textcolor{black}{By dividing the time series into different windows along time, where each channel within a window is treated as a node, and the correlations between channels are considered as edges, multivariate time series can be transformed into graph-based data.}
The GNN-based methods can be classified into dense and sparse graphs. In \textit{dense graphs}, each node is typically connected to almost all other nodes, with the edges often representing the strength of correlation or the probability of correlated influence. Methods based on dense graphs, such as GTS~\cite{gts} and FourierGNN~\cite{FourierGNN}, generally follow a CD strategy. In contrast, \textit{sparse graphs} only retain necessary edges, with most nodes remaining independent. For instance, MTGNN~\cite{MTGNN} preserves K edges per node, constructing a sparse K-regular graph. Different from this, MTSF-DG \cite{zhao2023multiple} sparsifies the adjacency matrix by filtering out low-probability edges based on a pre-set threshold. Methods based on sparse graphs belong to the CP strategy.
% 此外研究人员根据时序数据的不同表现，建立了不同类型的图，如图~\ref{GNN}所示，从不同的角度探索了GNN于时序预测的潜在贡献。

% \textbf{GNN-based:} The graph represents a specialized data structure that effectively describes the relationships between different entities. By treating each channel within a window as a node and the correlations between channels as edges, multivariate time series can be transformed into graph-based data. 
% The graph-based learning process allows for better exploration of correlations and message propagation, fully integrating feature and structural information to more efficiently handle dependencies between variables.
% 图是一个专门的数据结构，能够有效地描述不同实体之间的关系。通过将窗口内每个通道视为节点，通道间间的相关性视为边，多变量时间序列可以转换为图结构数据。基于图的学习过程可以更好的进行相关性的挖掘与消息的传递，充分整合特征与结构信息，更高效地处理变量之间的依赖关系。研究人员根据时序数据的不同表现，建立了不同类型的图，如图~\ref{GNN}所示，从不同的角度探索了GNN于时序预测的潜在贡献。
\begin{figure}[t!]
  \centering
    \includegraphics[width=1\linewidth]{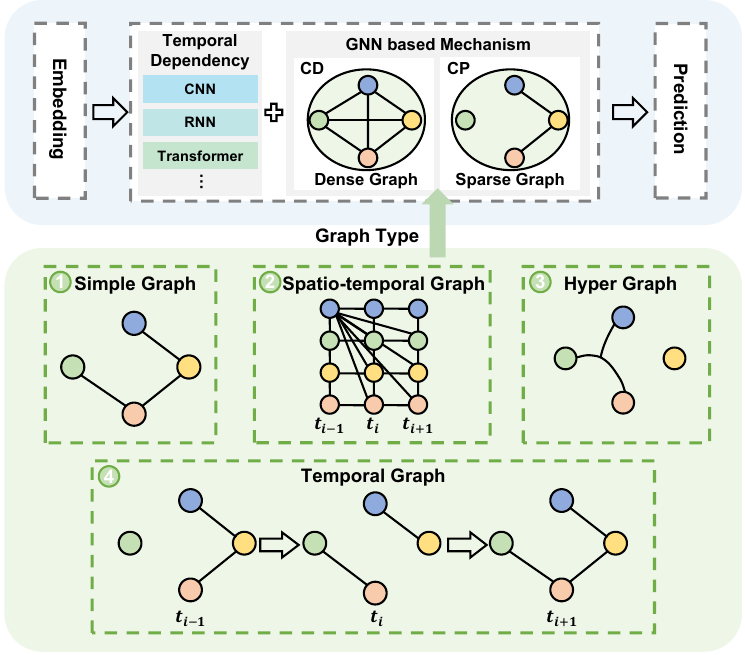}
  \caption{GNN-based mechanism for channel strategy.}
    \vspace{-4mm}
  \label{GNN}
\end{figure}
% Furthermore, as illustrated in Figure~\ref{GNN}, the explored GNN-based approaches can be broadly categorized as follows:
The sparsity of the constructed graph determines whether the method follows a CD or CP strategy. Additionally, GNN-based models often rely on the type of graph they construct when implementing the CD or CP strategy. As shown in Figure~\ref{GNN}, we classify graph types as follows:
% Beyond categorization based on sparsity, existing methods can also be distinguished by the types of nodes and edges they utilize, as illustrated in Figure~\ref{GNN}:}
% Researchers have established different types of graphs based on the varying characteristics of time series data, as shown in Figure~\ref{GNN}, exploring the potential contributions of Graph Neural Networks (GNN) in time series forecasting from different perspectives.
% 简单图是最基本的图模型，每一对节点之间最多只有一条边，需要定义良好的图结构进行消息传递。研究人员先后使用了通道相似度度量 (MTGNN,MSGNet,CrossGNN),数据相似度度量(GTS, WaveForM)来学习多变量间的相关性图结构。以时域(MTGNN,MSGNet,CrossGNN, GTS)或频域(WaveForM)信息作为节点学习特征。并在简单图中使用基于图卷积的消息传递，完成通道间依赖信息的传递。
I) \textbf{Simple Graph:} A simple graph is the most basic graph model, where there is at most one edge between each pair of nodes. A well-defined graph structure is required for effective message passing. Researchers have used channel similarity metrics (MTGNN, MSGNet~\cite{MSGNet}, CrossGNN~\cite{CrossGNN}) and data similarity metrics (GTS, WaveForM~\cite{WaveForM}) to learn the correlation graph structure among multivariate channels. They utilize time-domain (MTGNN, MSGNet, CrossGNN, GTS) or frequency-domain (WaveForM) information as node learning features. Graph convolution-based message passing is applied within the simple graph to facilitate the transmission of dependency information among channels.
%随着时空图模型在时空预测领域取得随着越来越多的成功，研究学者将其引入多变量时序预测之中，来解决时序模块与GNN的潜在不兼容，探索利用纯GNN的方式解决多变量时序问题,FourierGNN、FC-STGNN 把不同时间窗口，不同通道的序列融入一张图中，将多变量时序建模为时空图，在时间、通道两个维度进行消息传递。为避免由于图节点数过多在图构建、消息传播两个阶段的复杂计算，二者都使用了全连接的构图方式，并分别傅里叶域卷积算子与移动-池化卷积获得了O(nlog(n))的时间复杂度
% 与simple graph不同的是，Spatio-temporal Graph把多个时刻的不同通道同时纳入一张图中，来进一步考虑不同时刻间通道之间的关系。这样的好处是可以同时利用GNNs提取时序与通道间的依赖，解决时序模块与GNN的潜在不兼容问题。例如FourierGNN与FC-STGNN. 为避免由于图节点数过多在图构建、消息传播两个阶段的复杂计算，二者都使用了全连接的构图方式，并分别傅里叶域卷积算子与移动-池化卷积获得了O(nlog(n))的时间复杂度。
II) \textbf{Spatio-temporal Graph:}
\textcolor{black}{Unlike a simple graph, a Spatio-temporal Graph incorporates multiple channels at different time steps into a single graph, further considering the relationships between channels across different time steps. This approach allows GNNs to simultaneously model both temporal and channel dependencies, effectively addressing potential compatibility issues between the temporal module and the GNNs. The main challenge of Spatio-temporal Graph-based methods is to address the efficiency issues in the graph construction and message passing stages. For example, FourierGNN uses fully connected graph construction and employs Fourier domain convolution operators to achieve a time complexity of \(O(Nlog(N))\). Similarly, FC-STGNN~\cite{FC-STGNN} adopts the same graph construction method and employs moving-pooling convolution to achieve the same time complexity.}
 % \textcolor{black}{As the spatio-temporal graph model has achieved increasing success in the field of spatio-temporal prediction, researchers have introduced it into multivariate time series forecasting to address the potential incompatibility between temporal modules and GNNs. For example, FourierGNN and FC-STGNN~\cite{FC-STGNN} utilize pure GNN methods to solve multivariate time series problems.}
% They integrate sequences from different time windows and channels into a single graph, modeling multivariate time series as spatio-temporal graphs with message passing in both time and channel dimensions.
III) \textbf{Hyper Graph:}
% 超图是一种图的扩展，允许其中的超边连接多个顶点，可以建模更高阶的组交互。基于超图的模型认为变量间的交互不是pair-wise的，而是多个变量共同进行group-wise交互,因此基于超图的模型天生适合于构建CP策略。ReMo~\cite{ReMo},Ada-MSHyper分别构建了多视角与多尺度的超图，并通过设计超图上的消息传递机制，使得消息进行group-wise的传播。值得注意的是，二者分别使用不同的MLP与聚类约束促进组间异质性的表达。
Hypergraphs are an extension of graphs that allow hyperedges to connect multiple vertices, enabling the modeling of higher-order group interactions. Models based on hypergraphs assume that the interactions among channels are not pairwise but involve group-based interactions among multiple channels. \textcolor{black}{Therefore, hypergraph-based models are inherently suitable for constructing CP strategies.} ReMo~\cite{ReMo} constructs multi-view and multi-scale hypergraphs, respectively, and designs message passing mechanisms on these hypergraphs to enable group-wise message propagation. It is noteworthy that they use different MLPs or clustering constraints to promote the expression of heterogeneity among groups.
IV) \textbf{Temporal Graph:}
% 在真实的世界中，时序数据的相关性往往随时间变化，形成动态关系图，MTSF-DG、TPGNN分别使用动态图、多项式图来建模相关性的变化规律。MTSF-DG通过将历史关系图与未来关系图结合，利用记忆网络与逻辑符号学习历史相关性对于未来相关性的影响。TPGNN则是将相关性关系矩阵表示为具有时变系数的矩阵多项式，学习相关性的变化规律。根据二者稀疏性的不同，二者分别属于CD、CP策略。
In the real world, the correlation of time series data often changes over time, forming dynamic relational graphs. MTSF-DG and TPGNN~\cite{TPGNN} use dynamic graphs and polynomial graphs, respectively, to model the variation patterns of these correlations. \textcolor{black}{The CP model} MTSF-DG combines historical and future relational graphs, leveraging memory networks and logical symbol learning to capture the impact of historical correlations on future correlations. 
% TPGNN represents the correlation matrix as a matrix polynomial with time-varying coefficients, using a timestamp embedding mechanism to generate periodic embeddings in order to capture the periodic changes in dynamic dependencies, thus learning the variation patterns of correlations. 
\textcolor{black}{The CD model} TPGNN represents the correlation matrix as a matrix polynomial with time-varying coefficients to learn the evolving patterns of correlations.
 % Based on their differences in sparsity, the two methods belong to the CP and CD strategies, respectively.

% \textcolor{black}{It is worth mentioning that in dense graphs, each node is almost connected to every other node, so methods based on dense graphs typically adopt the CD strategy. In contrast, in sparse graphs, while there are necessary edges, most nodes remain independent, so methods based on sparse graphs typically adopt the CP strategy.}
% % 值得一提的是，稠密图中，每个节点与几乎所有节点之间都存在边，因此以稠密图为基础的方法往往属于CD策略，而稀疏图中，即存在必要的边，又使得大部分节点保持独立，因此以系数图为基础的方法属于CP策略。

\textbf{Others:} 
In addition to the mechanisms mentioned above, some models have proposed alternative approaches. For example: I) The CD model SOFTS~\cite{SOFTS} introduces the STAR module, which utilizes a centralized structure to first aggregate information from all channels using MLPs, and then distribute the aggregated information to each channel. This interaction not only reduces the complexity of inter-channel interactions but also minimizes reliance on individual channel quality. II) \textcolor{black}{The CP model LIFT~\cite{zhaorethinking} proposes a novel plugin, adaptable to all MTSF specific models, that efficiently estimates leading indicators and their lead steps at each time step. This approach enables lagging channels to utilize advanced information from a predefined set of leading indicators.} III) C-LoRA~\cite{C-LoRA} introduces a channel-aware low-rank adaptation (C-LoRA) plugin, which is adaptable to all MTSF-specific models. It first parameterizes each channel with a low-rank factorized adapter to enable individualized treatment. The specialized channel adaptation is then conditioned on the series information to form an identity-aware embedding. Additionally, cross-channel relational dependencies are captured by integrating a globally shared CD model.

% 特定模型的关键特点在于，它们要么忽略通道之间的依赖关系，要么专注于捕捉训练数据中通道之间的依赖，然后在测试数据上面应用。这些模型通常高度依赖于学习特定数据中各通道的关系，因此在面对新的或未见过的数据集时，它们的灵活性较差，这些模型的扩展性和适应性有限，难以有效应对新的数据集。

% \subsubsection{Multimodal Model}
% Multimodal models extend traditional time series forecasting by incorporating data from multiple modalities, such as text, images, or other sensor readings. By integrating diverse sources of information, these models can uncover richer patterns and relationships that are difficult to detect in single time series data.
% 多模态模型通过结合多种模态的数据（例如文本、图像或其他传感器读取数据）扩展了传统的时间序列预测。通过整合多样化的信息来源，这些模型可以发现单一时间序列数据中难以察觉的更丰富的模式和关系。

% \input{figs/taxonomy}

\begin{figure}[t!]
  \centering
    \includegraphics[width=1\linewidth]{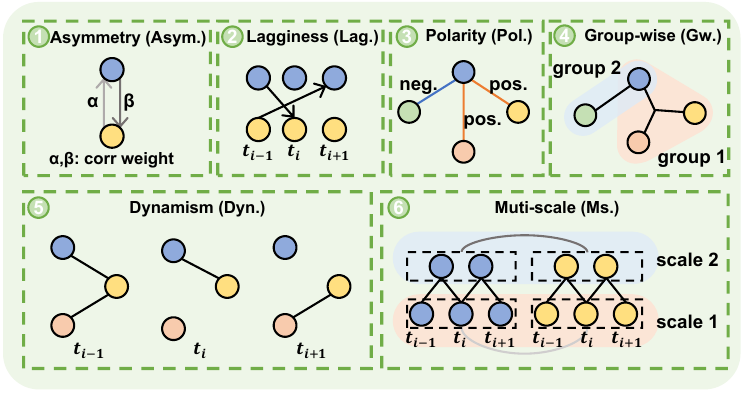}
  \caption{Characteristics perspective overview.}
    \vspace{-2mm}
  \label{Characteristics}
\end{figure}

\subsection{Characteristics Perspective}
\label{Characteristics Perspective}
To better explore the channel correlation in MTSF, it is often necessary to delve into the different characteristics of correlations among time-series channels. This section will explain the six key characteristics \textcolor{black}{(Figure~\ref{Characteristics})} commonly considered in current methods.

% 为了更好地探讨MTSF中的通道依赖关系，通常需要基于时间序列变量之间相关性的不同特征进行深入讨论。本节将阐述当前方法中常考虑的六个主要特征。
% \subsubsection{Asymmetry}
% \noindent
~~\textbf{Asymmetry:} 
% 非对称性是指在多变量的时间序列数据中，变量之间的相互关系并不是完全对等的，相互影响的程度不完全相同。基于transformer、MLP的方法由于其计算方式的特殊性天然具有非对称性，可以很好的表达相关性的非对称性。而基于GNN的方法则是通过构建非对称的距离度量建立有向的带权图，使对应的交互边在不同的传递方向上具有不同的权重，如MSGnet、GTS等。
Asymmetry refers to the unequal relationships among channels in multivariate time series, where the degree of mutual influence is not identical across channels. Methods based on transformers and MLP, due to the specific nature of their computational processes, inherently possess asymmetry, enabling them to effectively capture the asymmetric correlations. On the other hand, methods based on GNN establish directed, weighted graphs through asymmetric distance metrics, allowing interaction edges to have varying weights in different transmission directions, as seen in models such as MTGNN~\cite{MTGNN}, MSGnet~\cite{MSGNet}.

% \subsubsection{Lagginess}
% \noindent\\
\textbf{Lagginess:} 
% 延迟性是指某通道的当前状态不仅仅依赖于其余通道当前状态，还可能受到其余通道过去状态的影响，基于这种时延特性，VCformer在计算关注度矩阵时，考虑了通道间多步的时延的共同作用。而FourierGNN、FC-STGNN则是通过时空全连接图直接在不同通道、不同时刻的表征间进行消息传递。LEFT则是结合先验知识与神经网络预测滞后步长。通过考虑相关性的延迟性，方法可以了解到通道间更普遍的交互，进而获取到更好的表征与预测。
% Lagginess refers to the phenomenon in multivariate time series where the mutual influence between channels does not manifest immediately when a change occurs in one channel, but rather with a time delay.
% Lagginess refers to the fact that the current state of a certain channel not only depends on the current states of the other channels but may also be influenced by the past states of the other channels. 
Lagginess refers to the fact that the current state of a certain channel is influenced by the past states of the other channels. 
Based on the lagginess characteristic, VCformer~\cite{vcformer} incorporates the joint effects of multi-step delays among channels when calculating the attention matrix. In contrast, FourierGNN~\cite{FourierGNN} and FC-STGNN~\cite{FC-STGNN} directly perform message passing between representations across different channels and time steps using spatiotemporal fully connected graphs. LIFT~\cite{zhaorethinking}, on the other hand, combines prior knowledge with neural network predictions to estimate the lag step. 

% \subsubsection{Polarity}
% \noindent\\
\textbf{Polarity:} 
% Polarity是指通通道间的相互作用存在着正相关与负相关的区别，在建模时可以对二者进行区分以防止混淆。CrossGNN就是以符号图的方式，将相关性分为正相关、负相关、不相关三类，在消息传递时融合正向负向的信息交互，更好的捕获到了相关性之间的异质性。
% Polarity refers to the heterogeneous nature of interactions between channels, where the relationships can be either positive or negative.
Polarity refers to the distinction between positive and negative correlations in the interactions among channels. During modeling, it is important to distinguish between these two types of interactions to avoid confusion. CrossGNN~\cite{huang2023crossgnn} utilizes a sign graph approach, categorizing correlations into positive, negative, and neutral relationships. During message passing, it integrates both positive and negative information exchanges, thereby capturing the heterogeneity of correlations more effectively.

% \noindent\\
\textbf{Group-wise:} 
% \subsubsection{Group-wise}
% 组交互的意思是通道间的相关性存在分组现象，同组内相关性强，不同组间相关性弱，且不同组间的相关性存在差异。CM、DUET以聚类的方式对通道进行分组交互，ReMo、Ada-MSHyper通过超边建立了组内的消息传递。此外CM、ReMo对不同的组使用不同的MLP进行特征提取，Ada-MSHyper基于损失对超边进行约束，这些不同的方法都促进了不同组间差异的表达。
Group-wise refers to a phenomenon in which correlations among channels exhibit a grouping structure, characterized by strong correlations within the same group, weak correlations among different groups, and varying correlation dependencies across different groups. CCM~\cite{chen2024similarity} and DUET~\cite{qiu2025duet} use clustering techniques to group channels for interaction, while ReMo~\cite{ReMo} and Ada-MSHyper~\cite{Ada-MSHyper} establish intra-group message passing through hyperedges. Furthermore, CCM and ReMo apply different MLPs for feature extraction within different groups. This approach facilitates the expression of differences among different groups.

% \subsubsection{Dynamism}
% \noindent\\
\textbf{Dynamism:} 
% 多变量时间序列通道间的相关性在不同时间步有着不同表现，总体呈动态变化。首先，基于MLP的方法（如TimeMixer，TTM）在不同时间步权重保持不变无法表达动态性。
% 使用Transformer来考虑通道关系的方法通常采用通常采用series token或patches token，基于series token的方法不可表示动态性，如（iTransformer,DUET)。而基于patch Token的方法,如(CrossFormer)在不同的time step有不同的注意力分数，可建模动态性。
% 而在GNN中，只有图结构随序列时间变化而变化的方法可以建模动态性，如MSGNet在每个time step分别计算图结构。但上述建模动态性的方法只是在不同时间步考虑了不同的通道关系，而MSTF-DG、TPGNN、ESG 则更进一步的建模出通道关系在不同时间步的变化规律。例如，TPGNN使用时变系数的矩阵多项式来拟合不同时刻的通道关系。
% 上述建模动态性的方法只是在不同时刻考虑了不同的通道关系，而在此基础之上，MSTF-DG、TPGNN、ESG则认为，不同时刻间的通道关系存在直接的联系。例如，MSTF-DG会使用之前的通道关系来直接推测当前时刻的通道关系。
% The correlation among channels in multivariate time series exhibits different behaviors at different time steps, showing an overall dynamic change. 
The correlation among channels in multivariate time series is characterized by time-varying relationships, showing an overall dynamic change. 
First, methods based on MLP (such as TimeMixer~\cite{wang2024timemixer}, TTM~\cite{ttm}), where the weights remain constant across time steps, fail to capture dynamism. Methods that use a Transformer to consider channel correlation typically employ series tokens or patch tokens. Methods based on series tokens, such as iTransformer~\cite{liu2023itransformer} and DUET~\cite{qiu2025duet}, cannot capture dynamism. However, methods based on patch tokens, such as Crossformer~\cite{zhang2022crossformer}, assign different attention scores at different time patches, enabling the modeling of dynamism.
In GNNs, only approaches where the graph structure changes over time can capture dynamism, such as MSGNet~\cite{MSGNet}, which computes the graph structure at each timestep. However, the aforementioned methods for modeling dynamism only consider different channel relationships at different time steps. In contrast, MSTF-DG~\cite{zhao2023multiple}, TPGNN~\cite{TPGNN}, and ESG~\cite{ESG} propose that there is a direct connection between channel relationships across different time steps. For example, MSTF-DG uses previous channel relationships to directly infer the current channel relationships.
% The correlation between channels in multivariate time series data varies at different time steps, exhibiting dynamic changes over time. Due to the invariance of weights across different time steps in MLP-based methods (TimeMixer~\cite{wang2024timemixer}, TTM~\cite{ttm}) and the limitations of methods that do not partition the look-back window (iTransformer~\cite{liu2023itransformer}, DUET~\cite{qiu2025duet}), both are unable to capture dynamic behavior. In contrast, Transformer-based models and most GNN-based models can derive different correlation coefficients for different input data, thereby reflecting the dynamic nature of the relationships. Furthermore, MSTF-DG~\cite{zhao2023multiple}, TPGNN~\cite{TPGNN}, and ESG~\cite{ESG} take this a step further by considering the underlying patterns of correlation changes across different time steps.

% \noindent\\
\textbf{Multi-scale:} 
% \subsubsection{Muti-scale}
% Muti-scale指通道间的相关性在不同的尺度(如 时分秒）上具有不同的表现。MSGNet、CrossGNN、Ada-MSHype在不同的尺度间建立了不同的图结构用于描述不同尺度的相关性差别，并通过不同程度的交互实现了不同尺度相关性信息的融合。考虑相关性的多尺度异构性可以帮助模型更好的理解时序数据的多尺度特征，从而生成更好的预测结果。
Multi-scale refers to the phenomenon where the correlations among channels exhibit different behaviors at various time scales (such as hours, minutes, or seconds). MSGNet~\cite{MSGNet} and Ada-MSHyper~\cite{Ada-MSHyper} establish different graph structures across scales to describe the variations in correlation at different levels, and they achieve the fusion of correlation information at different scales through varying degrees of interaction. Considering the multi-scale heterogeneity of correlations helps the model better understand the multi-scale features of time series data, thereby generating more accurate predictions.

% \begin{table}[t!]
% \centering
% \caption{Comparison among different channel strategies.}
% \resizebox{0.9\columnwidth}{!}{
% \begin{tabular}{l|c|c|c}
% \toprule
% \textbf{Dimension}           & \textbf{CI}       & \textbf{CD}           & \textbf{CP}           \\ 
% \midrule
% \textbf{Efficiency}          & High              & Low                   & Moderate                     \\ \hline
% \textbf{Robustness}          & High          & Low              & Moderate   \\ \hline
% \textbf{Generalizability}    & Low               & Moderate              & High   \\ \hline
% \textbf{Capacity}            & Low               & High                  & Moderate                  \\ \hline
% \textbf{Ease of Implementation} & High           & Moderate                                   & Low                      \\ \bottomrule
% \end{tabular}}
% \label{tab:channel_comparison}
% \end{table}

\begin{figure}[t!]
  \centering
    \includegraphics[width=1\linewidth]{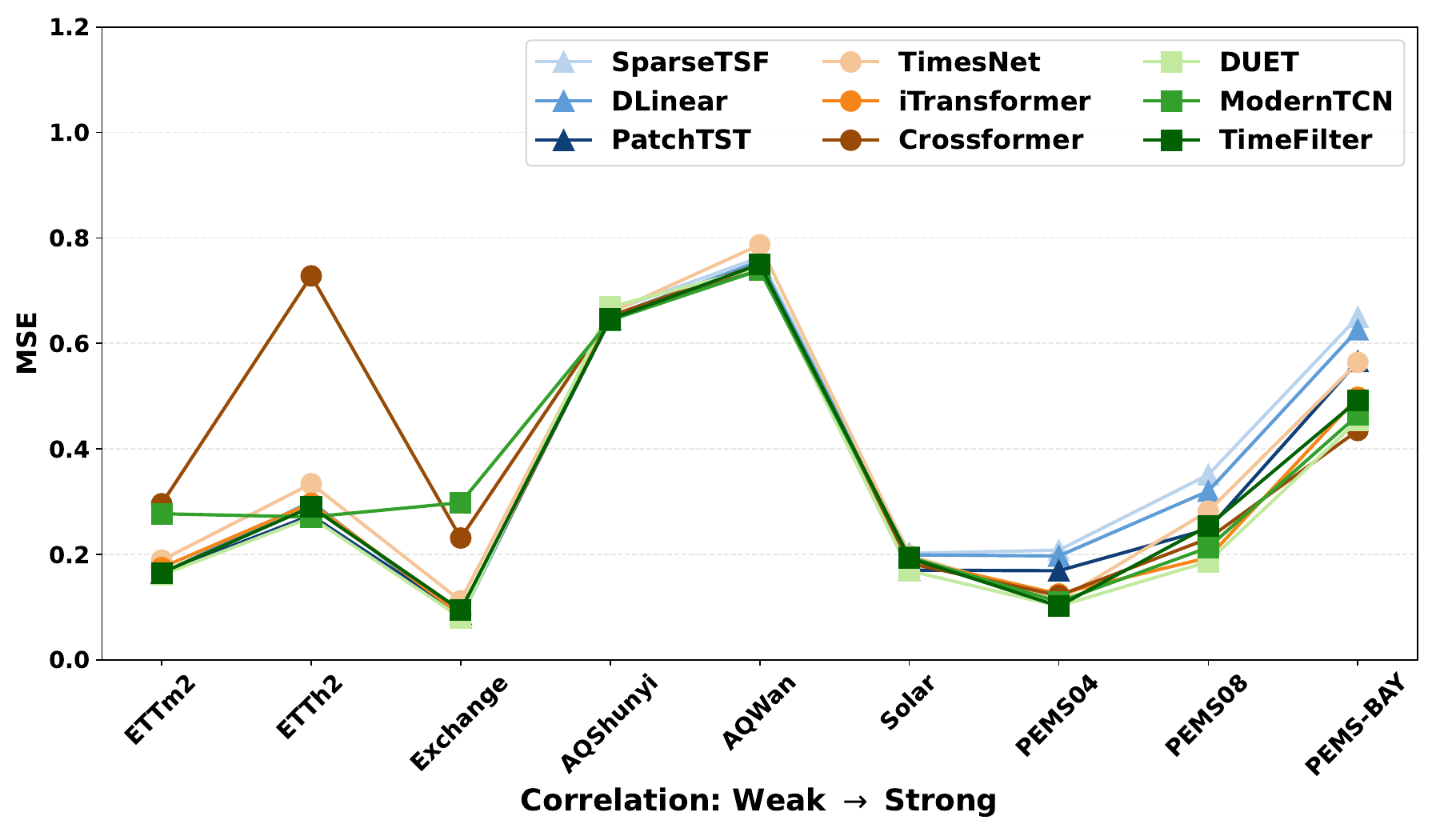}
  \caption{Method performance for varying correlation within datasets. Datasets are sorted by correlation (Weak → Strong).}
    \vspace{-2mm}
  \label{datasets_visual}
\end{figure}

\begin{figure}[t!]
  \centering
    \includegraphics[width=1\linewidth]{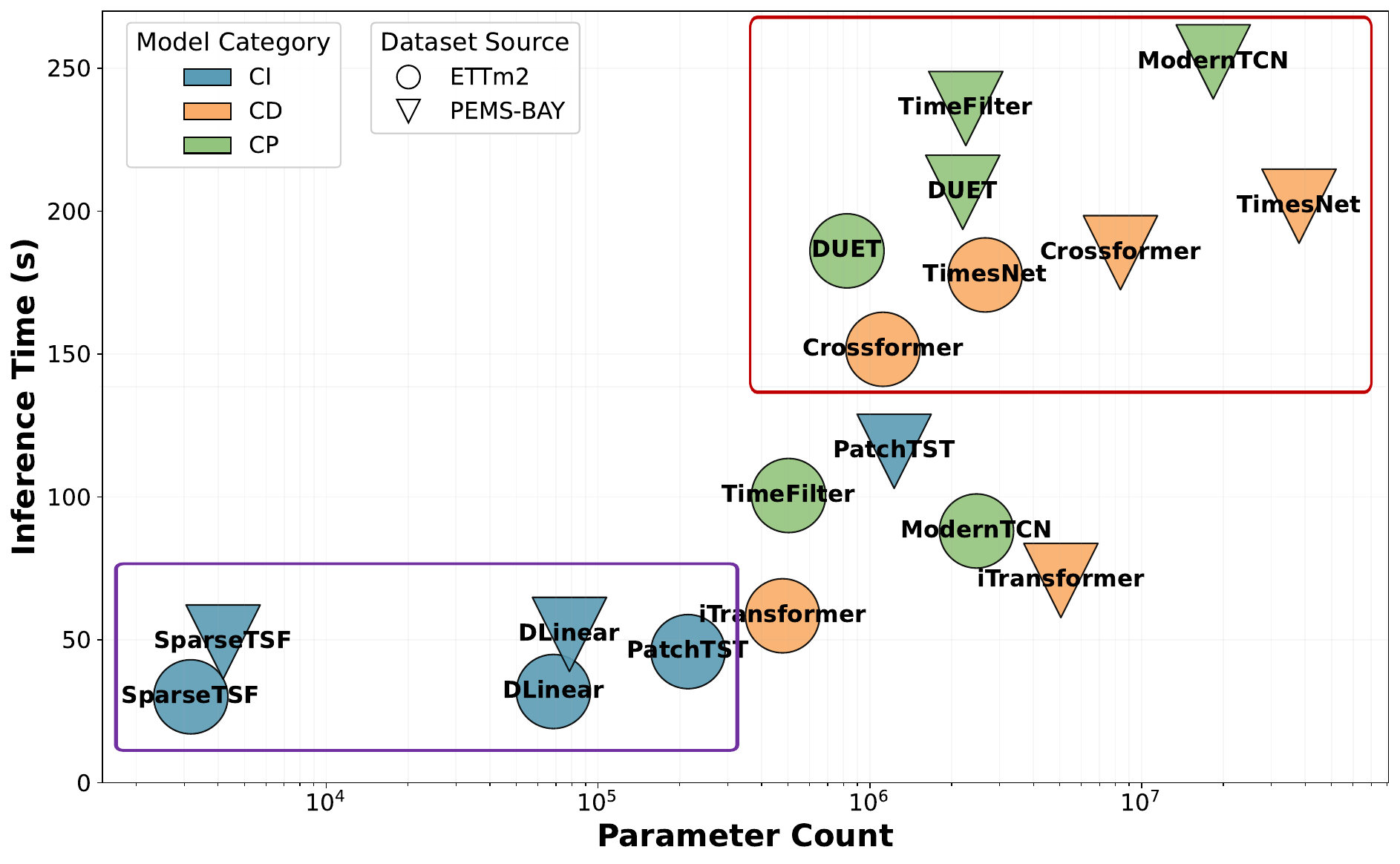}
  \caption{Comparison of parameter counts and inference time for deep learning methods.}
    \vspace{-2mm}
  \label{model_compare_visual}
\end{figure}

\section{Comparison within the Taxonomy}
\label{Comparison within the Taxonomy}
This section compares the advantages and limitations of CI, CD, and CP from multiple dimensions. We select three representative models for each category. To evaluate these three-channel strategies, we conduct two experiments.

% 在本节中，我们从多个维度比较了 CI、CD 和 CP 的优势与局限性。我们分别为每一个列别选起来3个具有代表性的模型。为了全面对比三种通道策略，我们进行了两项实验。在图~\ref{datasets_visual} 中，我们报告了在回看512，预测长度为 96 的设置下，从权威的时间序列预测benchmark，TFB中按照数据集的相关性强弱选取了十个数据集，报告均方误差 (MSE)。为了便于视觉区分，模型按类别进行了颜色编码：蓝色、橙色和绿色色调分别代表 CI、CD 和 CP 模型族。对于图~\ref{model_compare_visual} 中展示的效率分析，我们报告了在批量大小 (batch size) 为 1 时，回看512，预测长度为96的推理时间和参数量。在此次对比中，我们专门选择了 ETTm2 和 PEMS-BAY 这两个数据集，因为它们分别代表了具有最小和最大通道间相关性的数据集。

\textbf{Performance across Correlation:} Figure~\ref{datasets_visual} presents the MSE using a lookback window of 512 and a prediction horizon of 96. The experimental data is drawn from the time series forecasting benchmark, TFB~\cite{qiu2024tfb}, from which nine datasets are selected to represent varying correlation strengths. For visual clarity, model families are color-coded: CI (blue), CD (orange), and CP (green). In weakly correlated scenarios (left), CI achieves superior predictive performance by naturally disregarding inter-channel noise, whereas CD exhibits degraded performance due to its tendency to overfit spurious correlations. However, as correlations strengthen (right), CI encounters a capacity bottleneck, causing its performance to lag behind that of CD and CP strategies. In these regimes, both CD and CP excel, with the CP strategy demonstrating the strongest generalization. By flexibly modeling inter-channel dependencies, CP effectively captures complex interactions while filtering noise, yielding superior performance compared to pure CD strategies.

\textbf{Efficiency and Computational Complexity:} Figure~\ref{model_compare_visual} details the inference time and model parameter counts using a batch size of 1, a lookback window of 512, and a prediction horizon of 96. We specifically select the ETTm2 and PEMS-BAY datasets as representatives of scenarios with minimum and maximum inter-channel correlations, respectively. The CI strategy (purple boxed region) adopts a channel-independent paradigm, decoupling multivariate time series into univariate processes and enabling weight sharing across all channels. This design effectively eliminates the computational redundancy associated with inter-channel interactions, thereby minimizing parameter counts and significantly reducing inference latency to achieve superior computational efficiency. In stark contrast, CD and CP strategies (red boxed region), driven by the need to explicitly model complex inter-channel dependencies, inevitably incorporate additional architectural components (such as attention mechanisms or dynamic graph structures). This architectural complexity results in a substantial computational burden, manifested as significantly higher parameter counts and inference costs compared to the CI strategy.

\section{Future Research Opportunities}
% \subsection{Real-Time and Streaming Data}
% % 在许多应用中，例如金融预测或工业监控，实时预测至关重要。针对流式数据的通道间相关性提取及使用将显著的影响对未来的预测。可以通过考虑持续学习或增量学习方法展开对实时数据、流式数据中通道依赖的研究，以适应流数据，并在动态环境中平衡效率与准确性，最终推动通道依赖在实时数据中的应用。
% In many applications, such as financial forecasting or industrial monitoring, real-time predictions are crucial. The extraction and application of channel correlations significantly impact the accuracy of future predictions. Future research could explore continuous learning or incremental learning methods, focusing on channel strategy in real-time and streaming data, in order to adapt to the characteristics of streaming data and balance efficiency and accuracy in dynamic environments. This would ultimately promote the widespread application of channel strategy in real-time data.

\subsection{Channel Correlation in Future Horizon}
% 现有的研究中很少有方法讨论预测窗口的相关性关系，预测窗口的相关性直接影响了预测结果的质量，虽然已经有一些方法如PTGNN、Dy通过时序图的方法预测出未来窗口的相关性关系并加以应用，但是他们做的短步预测，且由于性能原因很难推广到长步预测。
Currently, few models address the correlation relationships within the prediction horizon. The correlation within the prediction horizon directly impacts the quality of the prediction results. Although some methods, such as TPGNN~\cite{TPGNN} and MTSF-DG~\cite{zhao2023multiple}, predict the channel correlation of the future horizon using temporal graph-based approaches and apply them accordingly, they focus on short-term forecasting and, due to performance limitations, are difficult to scale to long-term forecasting.

\textcolor{black}{
\subsection{Other Correlation Characteristics}
% 现有方法探索并分析了通道间相关性的非对称性、滞后性等6种特性对于预测性能的提升，而在真实场景中，相关性还隐含着更多特性，如：含噪性、条件性、多频性等等，探索更多的特性可以从机理上帮助模型更好的认知和利用变量间的相关性，并最终将其作用于预测与推理。
Existing research methods have explored and analyzed six characteristics of channel correlations---see section~\ref{Characteristics Perspective}. However, in real-world scenarios, correlations also contain additional characteristics, such as: I) \textbf{Multi-component:} DLinear~\cite{zeng2023transformers} and AutoFormer~\cite{wu2021autoformer} have demonstrated that decomposing time series into multiple components, such as trend and seasonality, significantly contributes to MTSF. Future research could explore how to model the channel correlations within each component separately, as well as how to integrate the channel correlations across multiple components. II) \textbf{Multi-frequency:} Correlations may manifest differently across various frequency components of time series data, and so on. Further exploration of these characteristics can help models better understand and utilize the correlations between channels, ultimately enhancing their predictive and inferential capabilities.}

% \subsection{How does the CHP dynamically select \(k\)?}
% In the Channel Hard Partiality (CHP) strategy, the number of channels associated with each channel, \(k\), is a critical factor influencing model performance and the effectiveness of channel dependency modeling. However, existing methods often rely on heuristics or manual tuning to determine \(k\), making them less adaptable to complex and dynamic data scenarios. Future research could explore data-driven methods for dynamically selecting \(k\) by analyzing inter-channel relationship matrices (e.g., correlation or distance matrices) to automatically determine the optimal number of clusters. 

% \subsection{Handling High-Dimensional Data}
% The traditional methods based on GNN and Transformer typically have a complexity of \(O(N^2)\), and existing datasets like Electricity and Traffic already have hundreds of channel dimensions, which significantly decreases the model's efficiency in these scenarios. As the number of channels in datasets continues to grow, model lightweighting has become an important challenge. Future research could explore how to efficiently process high-dimensional data while maintaining inter-channel relationships, for example, through dimensionality reduction, sparse modeling, or distributed computing techniques.

\subsection{Multi-modality for Channel Correlations}
Multiple modalities can be introduced to more comprehensively model the correlation among channels. Compared to a single time series modality, multimodal data can provide richer information sources, such as text, images, or other sensor data, which can compensate for potential gaps in time series data. By extracting features from multimodal data, the unique characteristics of channels across different modalities can be captured. Subsequently, cross-modal relational modeling mechanisms, such as cross-modal attention mechanisms or GNNs, can be employed to uncover the dynamic dependencies among channels. Additionally, to further enhance the modeling of channel correlation, an adaptive fusion mechanism can be designed to dynamically adjust interaction weights based on the correlations among different modalities. 
% 可以引入多种模态数据来更全面地建模变量之间的关联性。相比单一时间序列模态，多模态数据可以提供更加丰富的信息源，例如文本、图像或其他传感器数据，这些信息能够弥补时间序列数据中潜在信息的不足。通过对多模态数据进行特征提取，可以捕获不同模态下的变量特性，随后利用跨模态关系建模机制（如跨模态注意力机制或图神经网络）来揭示变量之间的动态依赖关系。此外，为了进一步提升对变量关联性的建模效果，可设计自适应融合机制，根据不同模态间的相关性动态调整交互权重，从而更加精准地刻画变量间的复杂关联。这种多模态方法有效利用了不同数据源的互补性，为变量关系的建模提供了更强的支持。

\subsection{Channel Strategy of Foundation Models}
Multivariate time series foundation models follow two main approaches: LLM-based models and time series pre-trained models. LLM-based models, lacking a channel dimension in language modality, typically adopt a CI strategy~\cite{time-llm}. Due to the high heterogeneity in the number of channels in time series data, most time series pre-trained models, such as Timer~\cite{Timer} 
% and Chronos~\cite{chronos}
, use a CI strategy to ensure robust predictions while avoiding complex channel correlation. In contrast, models like MOIRIA~\cite{MOIRAI} and UniTS~\cite{units} incorporate channel correlation during pretraining. MOIRIA flattens all channels, using positional embeddings to distinguish them, capturing both temporal and channel relationships with self-attention, while UniTS directly captures channel correlations via self-attention in the channel dimension. \textcolor{black}{However, time series pre-trained models only consider the CD strategy to capture the channel correlations, without fully considering the intricate and diverse channel correlations in different pre-training datasets, where the CP strategy may achieve better performance. There is also lacks works in LLM-based models to consider the channel correlations combined with the multimodal data.} Existing approaches remain relatively basic, leaving significant room for improving channel strategies in foundation models.

\section{Conclusion}
In this survey, we provide a comprehensive review of deep learning methods for MTSF from a channel strategy perspective. We categorize and summarize existing approaches using a proposed methodological taxonomy, providing a structured understanding of the field. Additionally, we offer insights into the strengths and limitations of various channel strategies and outline future research directions to further advance MTSF.

\bibliographystyle{named}
\bibliography{ijcai26}

\end{document}